\documentclass[runningheads,a4paper]{llncs}

\usepackage{amssymb}
\usepackage{amsmath}
\usepackage{graphicx}
\usepackage{subfig}
\usepackage{multirow}
\usepackage{wrapfig}
\usepackage{floatflt}
\usepackage{arydshln}
\usepackage{graphics}
\usepackage[lined,ruled,linesnumbered]{algorithm2e}
\usepackage[margin=2cm]{geometry}

\usepackage{url}
\urldef{\mailsa}\path|{a.agraharibaniya, sunil.aryal}@deakin.edu.au|
\urldef{\mailsb}\path|santosh.kc@usd.edu|

\begin{document}

\mainmatter  


\title{A Novel Data Pre-processing Technique: Making Data Mining Robust to Different Units and Scales of Measurement}

\titlerunning{ARES: Average Rank over an Ensemble of Sub-samples}
%
%

\author{Arbind Agrahari Baniya$^{1}$, Sunil Aryal$^{1}$ and Santosh KC$^{2}$}
\authorrunning{Agrahari Baniya et. al.}


\institute{$^1$Deakin University, Geelong, Victoria, Australia \\ \mailsa \\ $^2$University of South Dakota, Vermillion, SD, USA \\ \mailsb}
%
%

\maketitle


\begin{abstract}

Many existing data mining algorithms use feature values directly in their model, making them sensitive to units/scales used to measure/represent data. Pre-processing of data based on rank transformation has been suggested as a potential solution to overcome this issue. However, the resulting data after pre-processing with rank transformation is uniformly distributed, which may not be very useful in many data mining applications. In this paper, we present a better and effective alternative based on ranks over multiple sub-samples of data. We call the proposed pre-processing technique as ARES --- {\bf A}verage {\bf R}ank over an {\bf E}nsemble of {\bf S}ub-samples. Our empirical results of widely used data mining algorithms for classification and anomaly detection in a wide range of data sets suggest that ARES results in more consistent task specific outcome across various algorithms and data sets. In addition to this, it results in better or competitive outcome most of the time compared to the most widely used min-max normalisation and the traditional rank transformation. 


\end{abstract}


\section{Introduction}
\label{sec_intro}

In databases, data objects represent real-world entities defined by a set of selected features or properties. For example, people can be defined by their name, age, salary etc. and cars can be represented by model, price, fuel efficiency etc. Each data record/instance represents an instance of real-world entity defined by their values of the selected features. Features can have numeric or categorical values. In this paper, we consider the case where features have numeric values. Let $D$ be a collection of $N$ data instances $\{{\bf x}^{(i)}\}_{i=1}^N$. Each data instance ${\bf x}$ is represented as a vector of its values of $M$ features $\langle x_{1}, x_{2}, \cdots, x_{M}\rangle$ where $\forall_j\mbox{  }x_j \in \mathbb{R}$ and $\mathbb{R}$ is a real domain.

In real-world applications, numeric features are often measured or recorded in some units or scales \cite{scalesOfMeasurement_Stevens1946,PhDThesis_Fernando2017,PhDThesis_Aryal2017,usfAD_Aryal2018}. For example, annual income of people can be recorded in integer scale as $x=100,000$ or using a logarithmic scale of base 10 like $x'=5$. Sample variability can be measured in terms of standard deviation ($std$)  or variance ($var$) ($std=\sqrt{var}$ or $var={std}^2$). Similarly, fuel efficiency of cars can be measured in km/ltr as $x=9.0$ or ltr/100km as $x'=11.11$ where one is the inverse of the other. 

\begin{figure}[t]
\centering
\subfloat[Integer scale ($\times 10,000$)]{\includegraphics[width=0.33\textwidth]{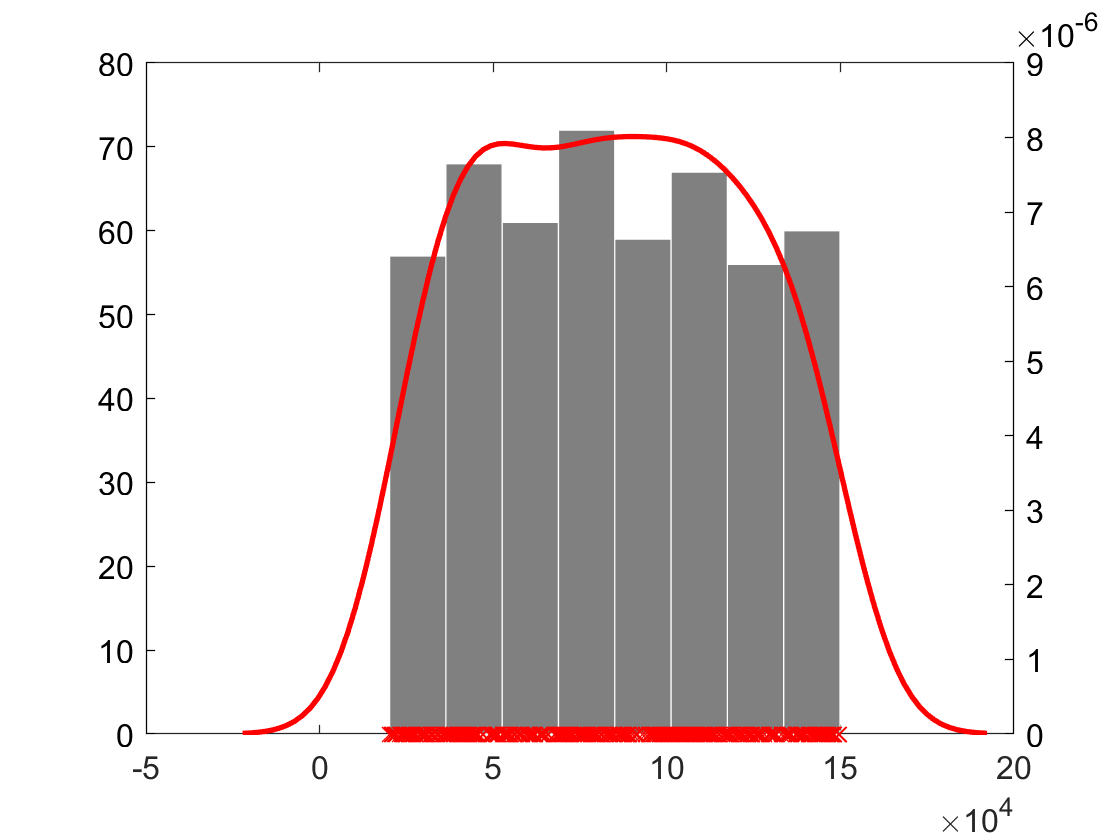}}
\hspace{50pt}
\subfloat[Log scale of base 10]{\includegraphics[width=0.33\textwidth]{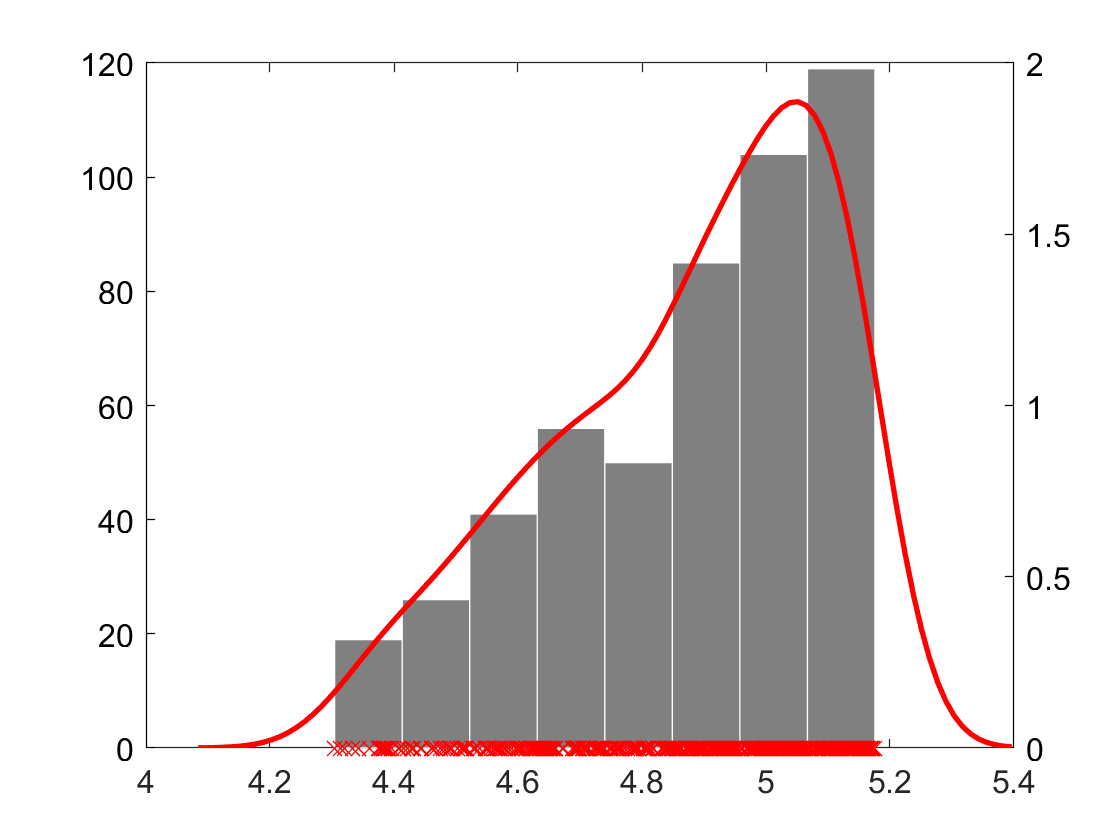}}
\caption{Distributions of annual incomes (in dollars) of 500 individuals in integer scale and logarithmic scale.}
\label{fig_example_income_data}
\end{figure}

Most of the existing data mining algorithms use feature values of data directly to build their models. They assume that data are embedded in an $M$-dimensional Euclidean space and use the spatial positions of data in the space. Many of them compute distances between data points and/or model density distribution of data. Because feature values depend on how features are measured, existing algorithms are sensitive to units and scales of measurement of features. They might give different patterns if data are measured in different units and/or scales. For example, Fig~\ref{fig_example_income_data} shows the distributions of annual incomes of 500 individuals in two scales - normal integer scale and logarithmic scale of base 10. The two distributions look very different from each other. Similarly, distances between data may also be changed if they are expressed/measured differently. For example, if $x=1, y=2$ and $z=3$, $y-x=z-y=1$. However, if the same data are provided as their squares ($x'=1,y'=4$ and $z'=9$), $y'-x'<z'-y'$.  

The task-specific performances of many existing data mining algorithms depend on how data are measured. They may perform poorly if data are not measured in appropriate units or scales. Unfortunately, information regarding units and scales of feature values are often not available during data mining, instead only the magnitudes of feature values are available. Even if known, the units and/or scales used for measurement may not be appropriate for the task at hand. Data need to be pre-processed to transform into an appropriate scale before using existing data mining algorithms. The simplest way to identify appropriate scale is to try different transformations and test which one produces the best task specific result. Because there is an infinite number of possible transformations, it is not feasible to find best transformation from trial and error \cite{SimUSF_Fernando2017}.

One simple solution is to use ranks of feature values. Rank transformation \cite{RankTransform_Conover1981} is robust to units and scales of measurement as it either preserves or reverses the ranks of data even though they are expressed differently. The main issue with the rank transformation is that it makes the distribution of resulting data uniform because the rank difference between two consecutive points is always 1 regardless of difference in the magnitude. Since the relative differences of data instances are lost, rank transformation may not be appropriate in some data mining applications.    

To overcome the above mentioned limitation of the rank transformation, we proposed a new variant of rank transformation using an ensemble approach. Instead of computing rank of data point $x$ among all $N$ points in $D$, we propose to compute the ranks of $x$ in $t$ sub-samples $D_j\subset D$ ($j=1,2,\cdots,t$) where $|D_j|=\psi<<N$ and use the average rank as its final transformed value. We call the proposed transformation technique  {\bf A}verage {\bf R}ank over an {\bf E}nsemble of {\bf S}ub-samples (ARES). Similar to  the rank transformation, ARES is also robust to units and scales of measurement. However, it also preserves the relative differences between data points to some extent.  

Our empirical results in the classification and anomaly detection tasks using different algorithms in a wide range of data sets show that pre-processing using ARES results in either better or competitive task specific results compared to widely used min-max normalisation and rank transformation based pre-processing of data. Furthermore, rank and ARES transformations are both robust to units and scales of measurement of raw data whereas min-max normalisation is not.

The rest of the paper is organised as follows. Section~\ref{sec_relatedWork} provides a brief review of previous works related to this paper. The proposed method is discussed in Section~\ref{sec_avensRank} followed by experimental results in Section~\ref{sec_exp} and conclusions in the last section.


\section{Related work}
\label{sec_relatedWork}

Psychologist S. S. Stevens (1946) \cite{scalesOfMeasurement_Stevens1946} discussed four types of scales of measurement and their statistical properties which have been largely argued by many researchers. Since then, the scales of measurement has been a subject of discussion mainly in the measurement theory and among psychology communities. Their argument is that raw numbers can be misleading as they could have been measured/presented in different ways. One should not conclude anything from a given set of numbers without understanding where they come from and the underlying process of generating them \cite{FootballNumbers_Lord1953,MeasurementScales_TownsendAshby1984,MeasurementScales_VellemanWilkinsom1993}. Although this issue is equally important if not more in automatic pattern analysis, it has been hardly discussed in the data mining literature.

Velleman and Wilkinson (1993) \cite{MeasurementScales_VellemanWilkinsom1993} argued that good data analysis does not assume data types and scales because data may not be what we see. Joiner (1981) \cite{LurkingVariables_Joiner1981} provided some examples of `lurking variables' --- data appear to have one type whereas in fact hide other information. However, in data mining, numeric data are assumed to be in `interval scale' \cite{PhDThesis_Fernando2017,SimUSF_Fernando2017} --- a unit difference has the same meaning irrespective of the values. This assumption may not be true when data are represented as a non-linear transformations, such as logarithm and inverse. 


Different features may have different ranges of values. For example, annual income of people can be in the range of tens of thousands to millions of dollars whereas age is in the range of 1 to 100 years. Since many data mining algorithms use feature values directly to solve different tasks, their analysis can be dominated by features with larger range of values. To address such dominance, data values in each features are re-scaled to be in unit range [0,1] \cite{PatternClassification_Duda2000} before passing them into data mining algorithms. This is referred to as min-max scaling or normalisation in the literature. Though it resolves the differences in the ranges of values across features resulted due to linear scaling (e.g., pounds vs kilograms or $^{\circ}$C vs $^{\circ}$F) where interval scale assumption holds, it does not resolve the differences in ranges resulted due to non-linear scaling of data (e.g., km/ltr and ltr/100km) where interval scale assumption is violated.

Recently researchers started to investigate the issue of units and scales of measurement in data mining \cite{PhDThesis_Fernando2017,PhDThesis_Aryal2017}. One potential solution suggested in the literature is to do rank transformation \cite{RankTransform_Conover1981} and use ranks instead of actual values. It is robust even to non-linear scaling because the rank is either preserved or reversed. Some prior work \cite{SimUSF_Fernando2017,MpKAIS_Aryal2017} show that rank transformation results in better task specific performances than using raw values with some data mining algorithms particularly those using distance/similarity of data instances. The main issue with the rank transformation is that it makes the distribution of resulting data uniform which can be detrimental in some data mining tasks.


Another line of research related to this area is representation learning \cite{RepresentationLearning_Bengio2013,RepresentationLearning_Zhong2016} where the task is to map data from the given input space into a latent space which maximises the task specific performance of a given data mining algorithm. Representation learning can be viewed as learning appropriate transformation that suits best for the data set and algorithm at hand. However, this approach has some issues: (i) requires extensive learning which can be computationally expensive in large and/or high-dimensional data sets; (ii) learns representation appropriate for the given algorithm, representation learned for one algorithm may not be appropriate for others in the same data set; and (iii) one cannot interpret the meaning of new features and what type of information they capture.   

Previous researches in this area are primarily focused on the algorithm level and developed new algorithms which are robust to units and scales of measurement \cite{PhDThesis_Fernando2017,PhDThesis_Aryal2017,SimUSF_Fernando2017,MpKAIS_Aryal2017,usfAD_Aryal2018}. In this approach, different set of algorithms need to be developed for each data mining task. We believe a better option would be to work in the data pre-processing level and develop a pre-processing technique robust to units and scales of measurement so that existing data mining algorithms can be used as they are. In this paper we propose one such technique to transform data such that the resulting distribution is similar even though raw data are given in different units and/or scales.


\section{The proposed method}
\label{sec_avensRank}

To address the issue of rank transformation resulting in uniform distribution of the transformed data, we propose a new variant of rank transformation using an ensemble approach. In each feature or dimension $i$, instead of computing rank of $x_i$ among all $N$ values, we propose to aggregate ranks of $x_i$ in $t$ sub-samples of values in dimension $i$. We call the proposed method as ARES ({\bf A}verage {\bf R}ank over an {\bf E}nsemble of {\bf S}ub-samples) transformation.

To make the explanation simple, we assume $D$ is a one-dimensional data set where $|D|=N$. We use $t$ sub-samples $D_j\subset D$ ($j=1,2,\cdots,t$) where $|D_j|=\psi<<N$. The transformed value of $x$, $\tilde{x}_{ARES}$, is the average rank of $x$ over all $t$ sub-samples.
\begin{equation}
    \tilde{x}_{ARES} = \frac{1}{t} \sum_{j=1}^t r(x|D_j)
    \label{eqn_ares}
\end{equation}
where $r(x|D_j)$ is the rank of $x$ in $D_j$:
\begin{equation}
    r(x|D_j) = |\{y\in D_j: y<x\}|
    \label{eqn_subsamplesRank}
\end{equation}

Based on the definition of $r(x|D_j)$ given by Eqn~\ref{eqn_subsamplesRank}, all $x\in D$ have ranks in $\{0,1,\cdots,\psi\}$ in $D_j$. If $s_j^{(1)},s_j^{(2)},\cdots,s_j^{(\psi)}$ are sorted samples in $D_j$:
  \begin{equation}
    r(x|D_j) =
    \begin{cases}
      0 & \text{if}\ x<s_j^{(1)} \\
      k & \text{if}\ s_j^{(k)}\leq x<s_j^{(k+1)} \text{and } 1\leq k \leq \psi-1\\
      \psi & \text{if}\ x\geq s_j^{(\psi)}
    \end{cases}
  \end{equation}


\begin{figure}[t]
\centering
\subfloat[$x$]{\includegraphics[width=0.3\textwidth]{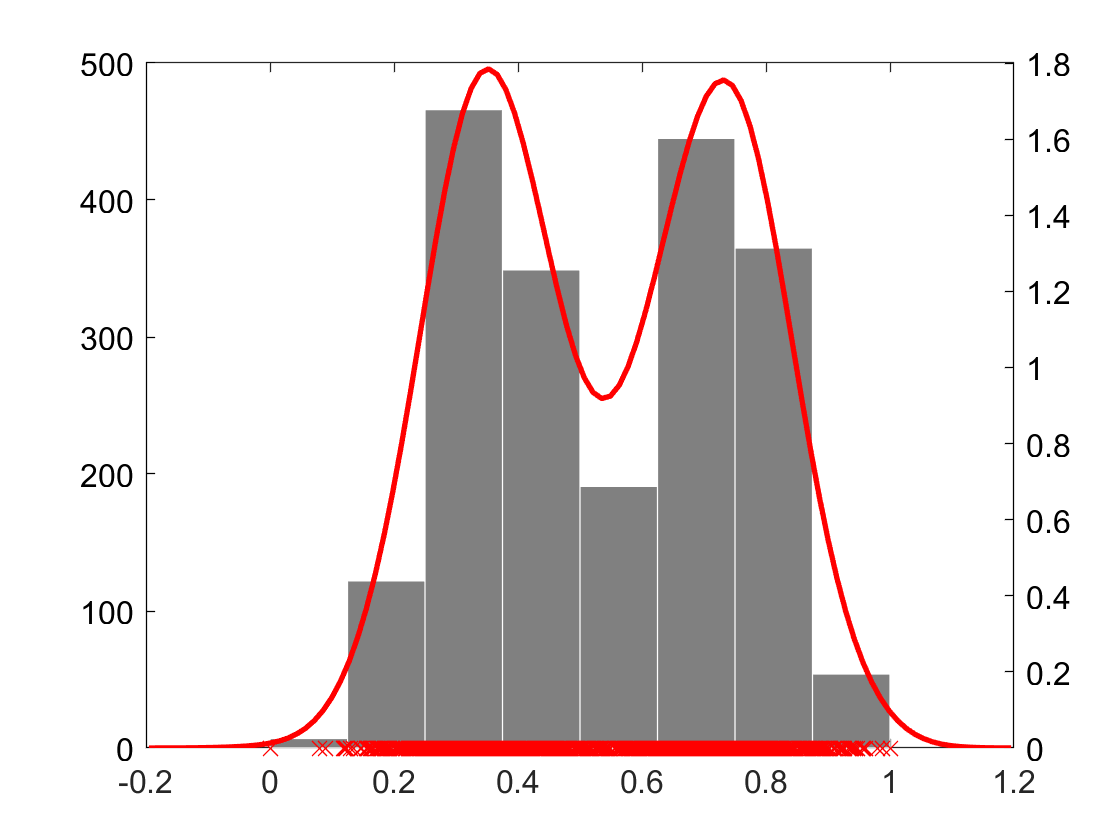}}
\subfloat[rank($x$)]{\includegraphics[width=0.3\textwidth]{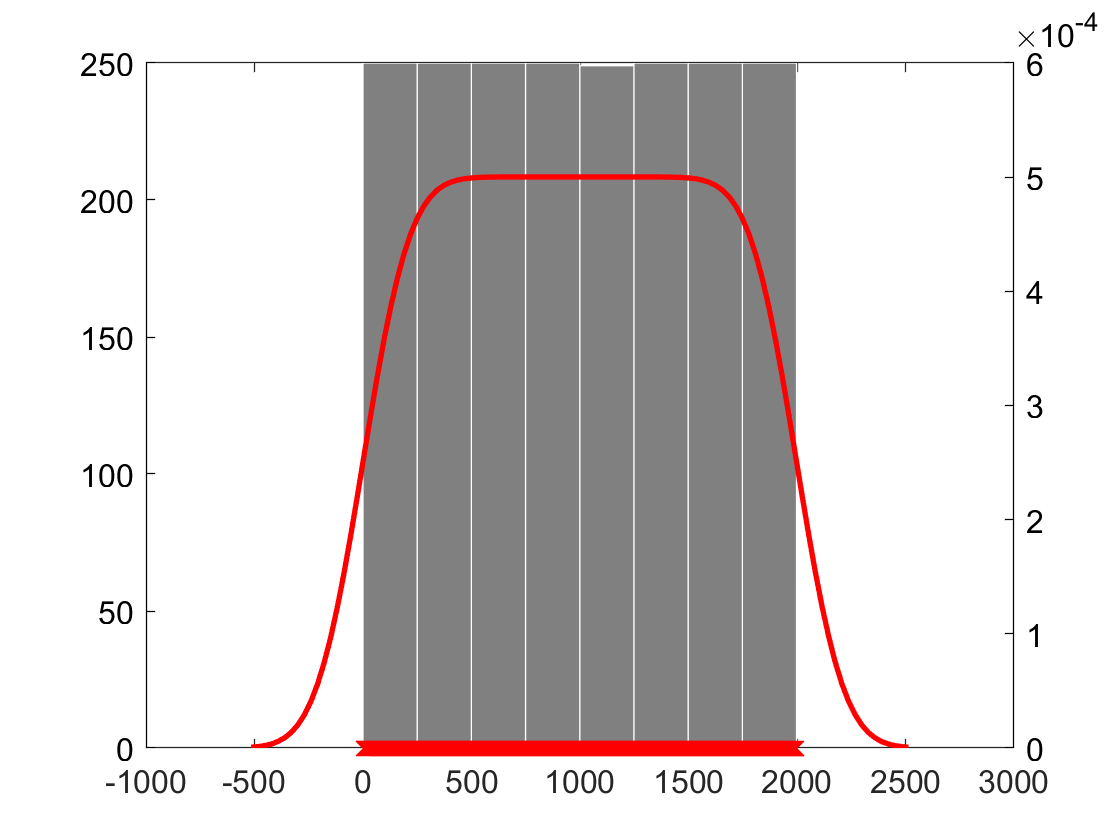}}
\subfloat[ares($x$)]{\includegraphics[width=0.3\textwidth]{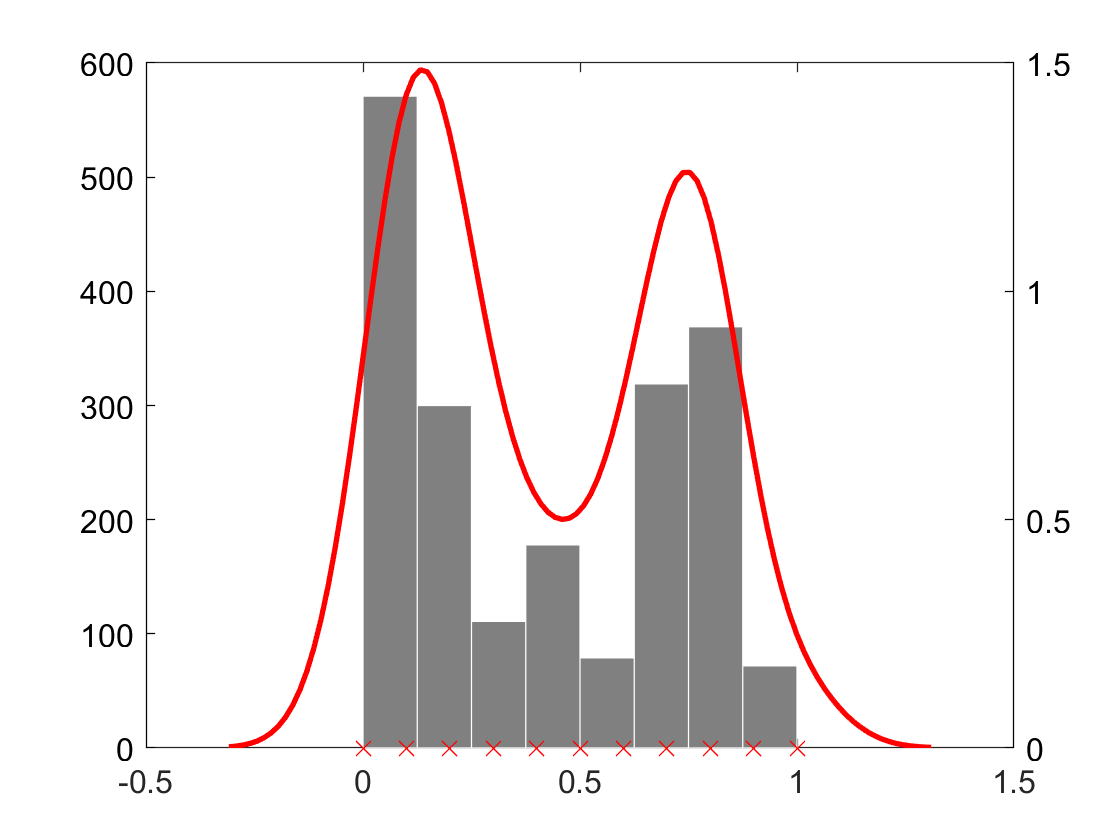}}
\\
\subfloat[log($x$)]{\includegraphics[width=0.3\textwidth]{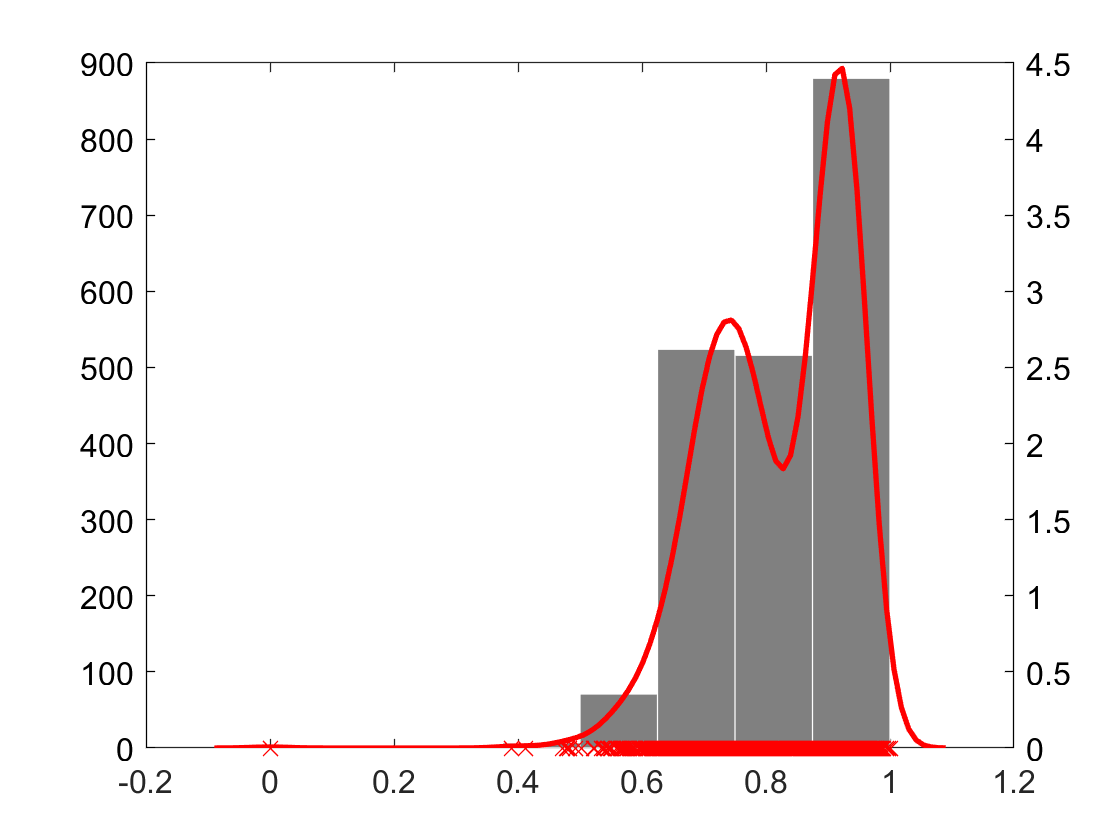}}
\subfloat[rank(log($x$))]{\includegraphics[width=0.3\textwidth]{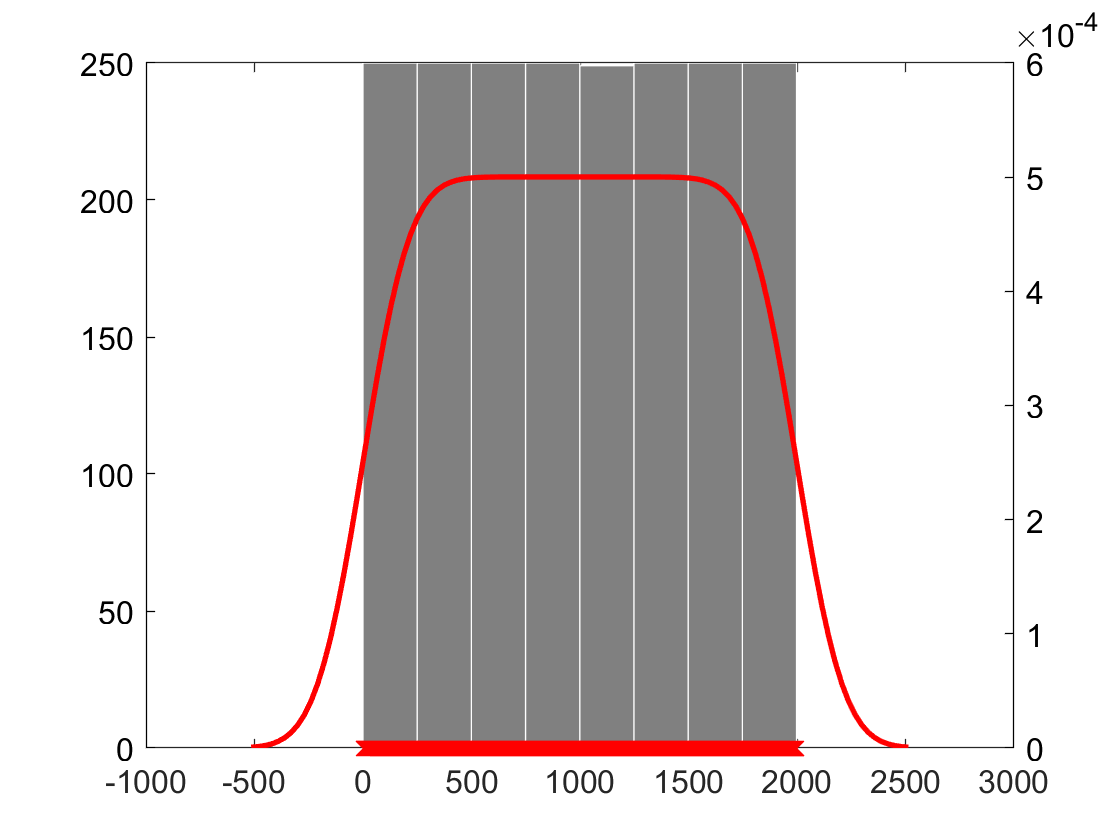}}
\subfloat[ares(log($x$))]{\includegraphics[width=0.3\textwidth]{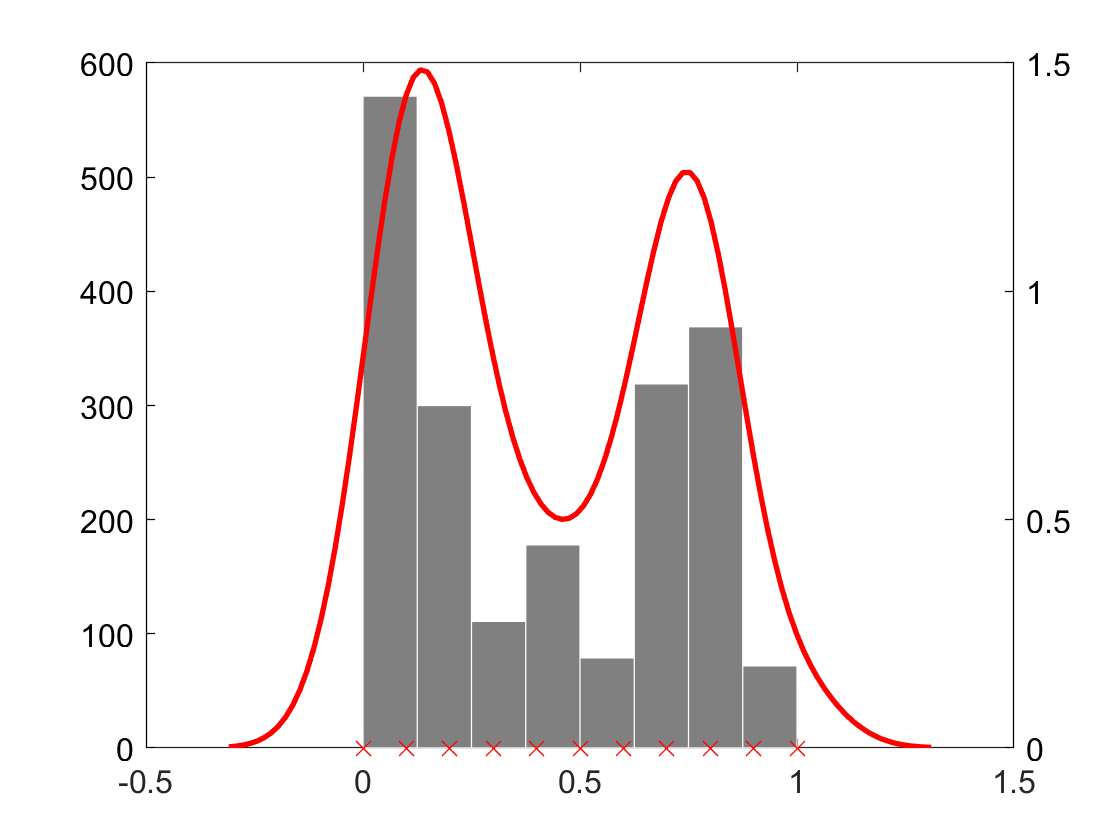}}
\\
\subfloat[inv($x$)]{\includegraphics[width=0.3\textwidth]{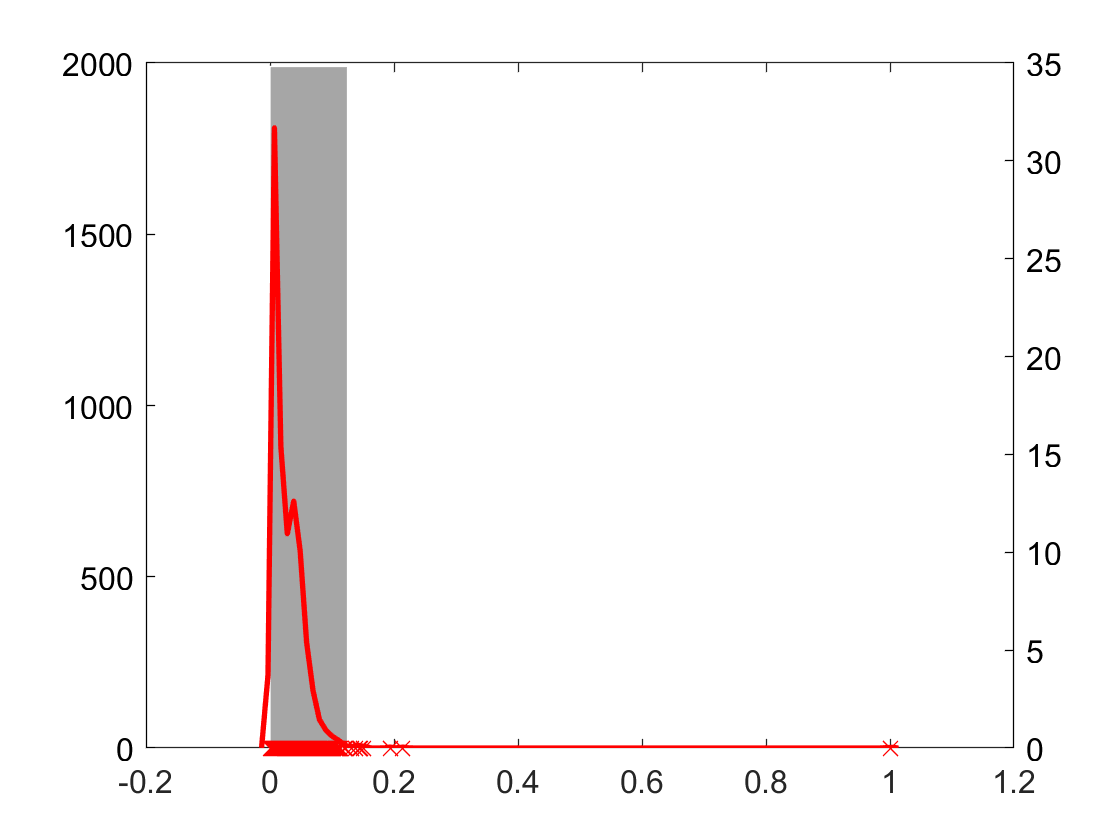}}
\subfloat[rank(inv($x$))]{\includegraphics[width=0.3\textwidth]{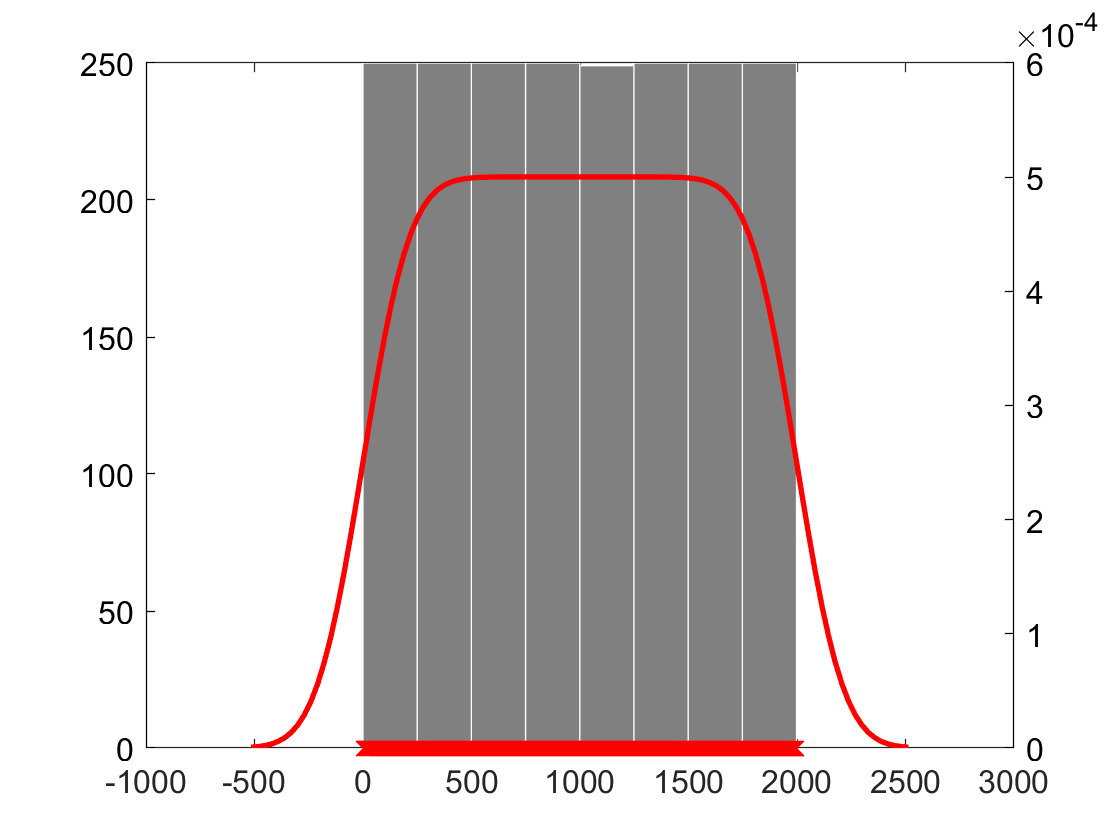}}
\subfloat[ares(inv($x$))]{\includegraphics[width=0.3\textwidth]{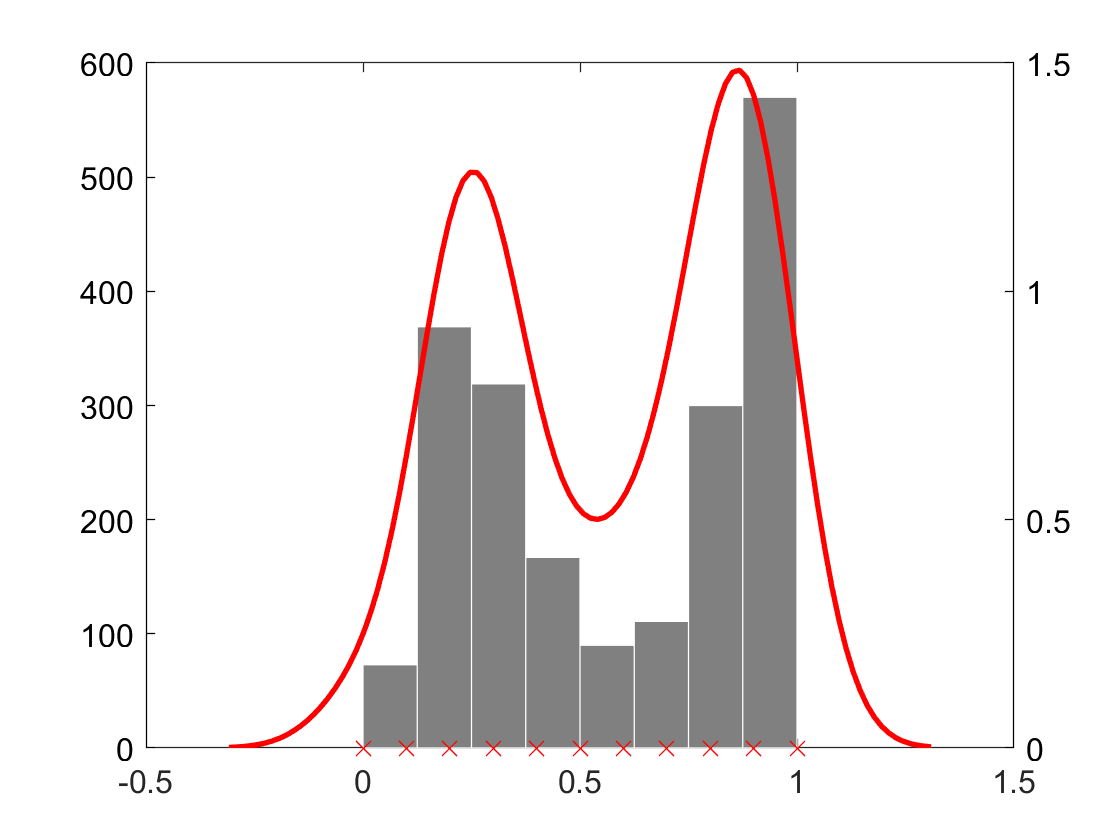}}
\caption{Distributions of an example data set and its logarithmic and inverse transformations and their pre-processing based on Rank and ARES. Histograms are plotted using the left y-axis and probability density functions are plotted using the right y-axis
}
\label{fig_example_rank_ares}
\end{figure} 

All $x\in D$ such that $s_j^{(k)}\leq x<s_j^{(k+1)}$ will get the same $r(x|D_j)$ of $k$ regardless of the their differences in magnitudes. They can not be differentiated. However, some of them will have different ranks in other sub-samples. Thus, averaging over different sub-samples will maintain their differences to some extent. 
For example, even the case of $|D_j|=\psi=1$ where all $x\in D$ have ranks either 0 or 1 depending upon either they lie on the left or right of the sample selected in $D_j$, doing it multiple times, say $t=10$, maintains the differences between data points to some extent. Fig~\ref{fig_example_rank_ares}(a-c) show the distribution of an example data set and its transformations using traditional rank and ARES. It is clear that ARES (Sub-figure c) preserves the differences in data values in the original distribution (Sub-figure a) better than the rank transformation using the entire $D$ (Sub-figure b).    

Like the traditional rank transformation, ARES is also robust to the changes in the scales and units of measurement. It is a variant of rank transformation using sub-samples in ensemble. Fig~\ref{fig_example_rank_ares}(d-i) show the same data as in Sub-figure (a) in logarithmic\footnote{The default base of logarithm in this paper is $e$, i.e., natural logarithm, unless specified otherwise} and inverse scales and their transformations based on the traditional rank and ARES. From Sub-figures c, f and i, it is clear that ARES results in the same distribution. Note that the resulting distribution in the case of inverse scale (Sub-figure i) is the reverse of that in the original and logarithmic scales (Sub-figure c and f). There is some small differences in the ARES transformations of samples selected in $D_j$ (i.e., $x=s_j^{(k)}$) in the case of inverse scale because of the `$<$' sign in Eqn~\ref{eqn_subsamplesRank}.   

In terms of run-time, ARES requires $O(t\psi\log\psi)$ time to generate and sort $\psi$ samples $t$ times and $O(Nt\log\psi)$ to compute ranks of all $N$ instances in $D$. The traditional rank transformation requires $O(N\log N)$ time to sort and rank $N$ instances in $D$.


\section{Empirical evaluation}
\label{sec_exp}

In this section, we present the results of experiments conducted to evaluate the effectiveness of ARES against the widely used min-max normalisation and the traditional rank transformation. We evaluated them in terms of their task specific performances and robustness to the changes in the units/scales of measurement in the classification and anomaly detection tasks using a wide range of publicly available data sets. 

In order to mimic the real-world scenario of possible variations in units/scales in data, monotonic transformations using logarithm, square, square root and inverse as $\log x, x^2, \sqrt{x}$ and $\frac{1}{x}$ were applied to data in each dimension. To cater for $x<0$ where $\log x$ and $\sqrt{x}$ are not defined, all transformations were applied after rescaling feature values in the range of [0,1] in all dimensions. Further, to cater for $x=0$ in the case of $\log x$ and $\frac{1}{x}$, all monotonic transformations were applied on $x'=b(x+a)$ where $a=0.0001$ and $b=100$. Scaling using $b=100$ was used to change the inter-point distances significantly. We used the same settings of monotonic transformation as discussed in \cite{SimUSF_Fernando2017}.

The proposed method ARES and the traditional rank transformation were implemented in Python using the Scikit Learn library \cite{scikit-learn_2011} which has the implementation of the min-max normalisation. Two parameters $\psi$ and $t$ in ARES were set empirically as default to 7 and 10, respectively. 

We discuss the experimental setups and results in the classification and anomaly detection tasks separately in the following two subsections. 

\subsection{Classification}

We compared the classification accuracies of the three contending pre-processing techniques (min-max normalisation, rank transformation and ARES transformation) in three widely used classification algorithms --- Artificial Neural Network (ANN) \cite{ANN_Rumelhart1988}, K-Nearest Neighbours (KNN) \cite{knn_Cover1967} and Logistic Regression (LR) \cite{GLM_McCullagh1989}. We used 15 publicly available data sets whose characteristics are provided in Table~\ref{tbl_classify_datasets}.

\begin{table}[t]
\scriptsize
\centering
\caption{Characteristics of data sets used in the classification task.}
\begin{tabular}{l @{\hspace{15pt}} r @{\hspace{15pt}} r @{\hspace{15pt}} r}
\hline
Name & \#Features ($M$) & \#Instances ($N$)  & \#Classes ($C$) \\
\hline
Churn & 13 & 10000 & 2  \\
Corel & 68 & 10000 & 100  \\
Diabetes & 9 & 768 & 2  \\
Glass & 10 & 214 & 6  \\
Heart & 14 & 303 & 2  \\
Miniboone & 51 & 129596 & 2  \\
Occupancy & 6 & 9752 & 2  \\ 
OpticalDigits & 63 & 5620 & 10  \\ 
Pageblocks & 11 & 5473 & 5  \\
Pendigits & 17 & 10992 & 10  \\
RobotNavigation & 25 & 5456 & 4  \\
SatImage & 37 & 6435 & 6   \\
SocialNetworkAds & 5 & 400 & 2  \\
Steelfaults & 26 & 1941 & 7  \\
USTaxPayer & 10 & 1004 & 3  \\
\hline
\end{tabular}
\label{tbl_classify_datasets}
\end{table} 

We conducted all classification experiments using a 10-fold cross-validation where 9 folds were used as the training set and the remaining one fold was used as the test set. The same process was repeated 10 times using each of the 10 folds as the test set. We reported the average classification accuracy over 10 runs. 

We used the Scikit Learn implementations of all three classifiers with default parameters settings except solvers in ANN and LR which were set to `\textit{lbfgs}' and `\textit{saga}', respectively, and `\textit{multi-class}' in LR which was set to \textit{multinomial}, to suit the multi-class classification problems. To make the results reproducible we set the `\textit{random state}' to 0 where applicable.

\begin{figure}[t]
\centering
\subfloat[ANN]{\includegraphics[width=0.33\textwidth]{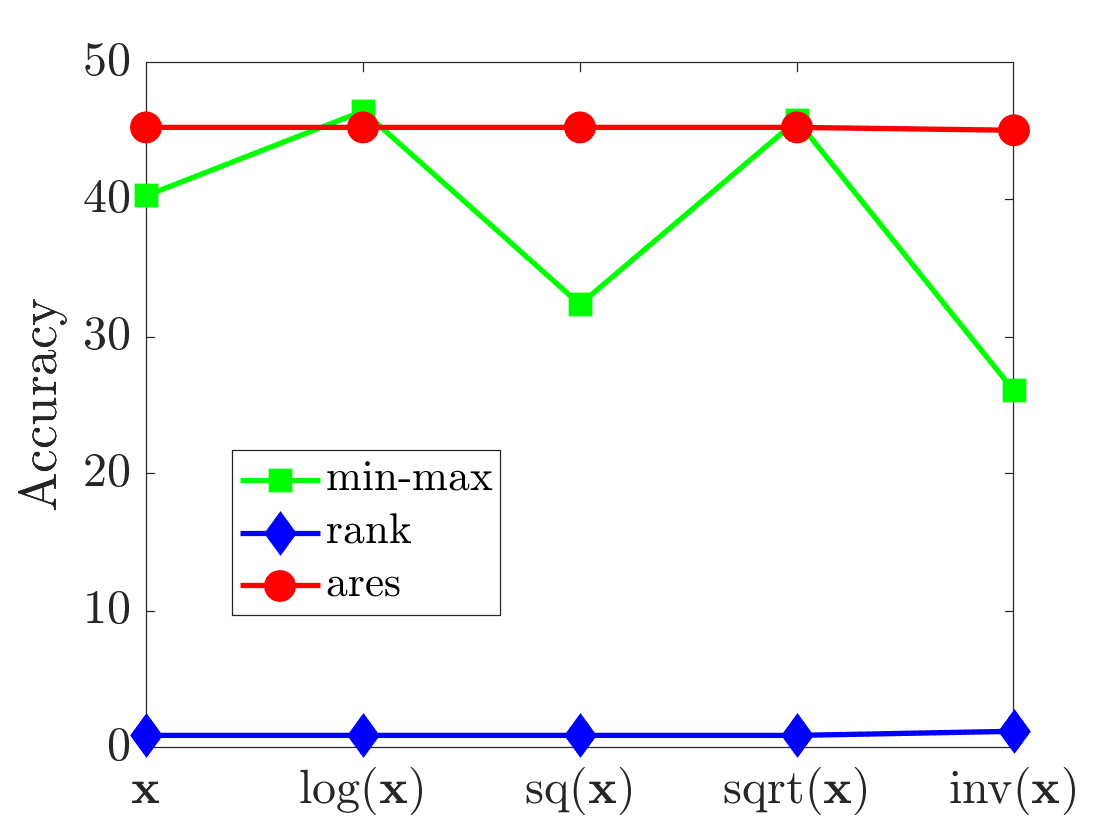}}
\subfloat[KNN]{\includegraphics[width=0.33\textwidth]{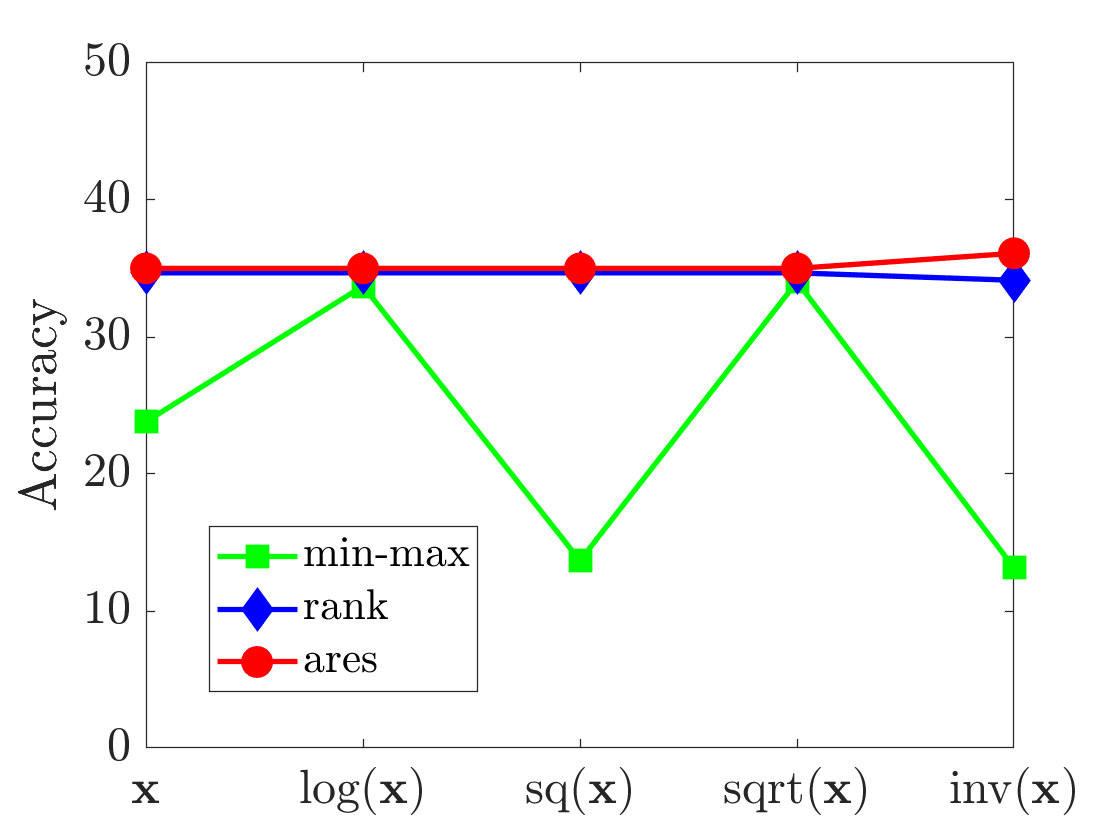}}
\subfloat[LR]{\includegraphics[width=0.33\textwidth]{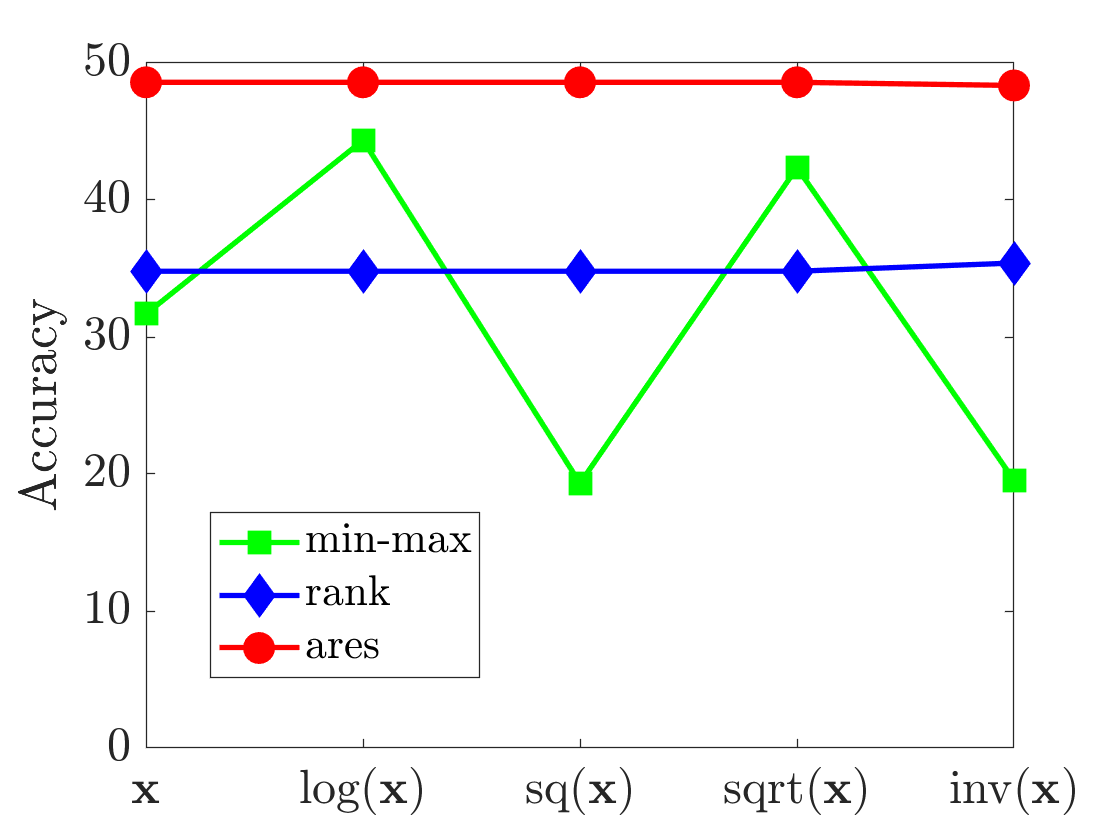}}
\caption{Average classification accuracy over a 10-fold cross validation in the Corel data set.
}
\label{fig_classify_corel}
\end{figure}

Because rank and ARES transformations produce the same results with all four monotonic transformations (there will be some small variations with the inverse transformation due to the issue discussed in Section~\ref{sec_avensRank}), we present the results of rank and ARES with the original input scale only. As an example to justify this, the detailed results of three contending pre-processing techniques with the original scale and the four monotonic transformations in the Corel data set is provided in Fig~\ref{fig_classify_corel}. The average classification accuracies of all three classifiers using three contending pre-processing techniques - min-max (original scale and its four monotonic transformations), rank (original scale) and ARES (original scale) are provided in Table~\ref{tbl_classify_accuracy}. 

\begin{table}[!htb]
\scriptsize
\centering
\caption{Average classification accuracy over a 10-fold cross validation. Overall best accuracy in each row is bold faced and the best accuracy result among monotonic transformations in the case of min-max normalisation is italicised.}
\begin{tabular}{c | l |  c @{\hspace{10pt}} c @{\hspace{10pt}} c @{\hspace{10pt}} c @{\hspace{10pt}} c @{\hspace{10pt}} | c @{\hspace{5pt}} | c }
\hline
\multirow{2}{*}{$\mbox{\hspace{15pt}}$} & \multirow{2}{*}{Data set} & \multicolumn{5}{c|}{min-max} & rank & ares \\
\cline{3-7}
&  & ${\bf x}$ & log(${\bf x}$) & sq(${\bf x}$) & sqrt(${\bf x}$) & inv(${\bf x}$) & ${\bf x}$ & ${\bf x}$ \\
\hline
\multirow{17}{*}{\rotatebox[origin=c]{90}{\parbox[c]{3.2cm}{Artificial Neural Network (ANN)}}} &  Churn &  85.48 &  85.80 &  85.62 &  \textbf{\textit{85.81}} &  85.61 &  76.98 &  84.57 \\
 &  Corel &  40.31 &  \textbf{\textit{46.49}} &  32.38 &  45.82 &  26.06 &  00.92 &  45.30 \\
 &  Diabetes &  \textbf{\textit{76.57}} &  74.74 &  74.62 &  76.44 &  67.19 &  59.50 &  74.09 \\
 &  Glass &  70.53 &  66.83 &  \textit{70.57} &  70.65 &  68.46 &  58.14 &  \textbf{70.80} \\
 &  Heart &  \textbf{\textit{79.59}} &  74.38 &  77.95 &  79.57 &  72.99 &  67.58 &  79.57 \\
 &  Miniboone &  \textit{92.38} &  90.20 &  91.89 &  92.22 &  84.34 &  71.84 &  \textbf{94.35}\\
 &  Occupancy &  99.27 &  99.04 &  \textbf{\textit{99.34}} &  99.27 &  96.20 &  72.74 &  99.18 \\
 &  OpticalDigits &  \textbf{\textit{97.67}} &  95.18 &  97.31 &  97.49 &  91.21 &  12.94 &  97.01 \\
 &  Pageblocks &  96.80 &  \textbf{\textit{97.30}} &  94.35 &  97.24 &  96.44 &  89.77 &  97.13 \\
 &  Pendigits &  \textbf{\textit{99.20}} &  93.71 &  99.07 &  99.02 &  66.68 &  10.58 &  99.01 \\
 &  RobotNavigation &  93.73 &  89.96 &  88.25 &  \textit{95.33} &  75.13 &  40.32 &  \textbf{\textit{96.70}} \\
 &  SatImage &  89.40 &  82.52 &  \textit{90.04} &  86.08 &  64.15 &  41.34 &  \textbf{90.07} \\
 &  SocialNetworkAds &  \textbf{\textit{89.75}} &  89.74 &  87.26 &  89.25 &  88.79 &  64.26 &  88.49 \\
 &  Steelfaults &  72.34 &  69.52 &  70.41 &  \textit{72.97} &  58.39 &  34.11 &  \textbf{73.17} \\
 &  USTaxPayer &  32.75 &  \textbf{\textit{36.17}} &  34.76 &  32.87 &  33.88 &  32.96 &  33.47 \\
\hline
\multirow{17}{*}{\rotatebox[origin=c]{90}{\parbox[c]{3.6cm}{K-Nearest Neighbours (KNN)}}} &  Churn &  80.10 &  \textbf{\textit{82.27}} &  79.81 &  80.79 &  82.20 &  76.79 &  \textbf{82.27} \\
 &  Corel &  23.82 &  33.72 &  13.68 &  \textit{34.04} &  13.15 &  34.68 &  \textbf{35.01} \\
 &  Diabetes &  73.70 &  66.15 &  \textbf{\textit{74.09}} &  70.32 &  63.81 &  73.58 &  73.69 \\
 &  Glass &  69.79 &  \textbf{\textit{70.61}} &  69.59 &  67.55 &  70.22 &  68.25 &  69.36 \\
 &  Heart &  81.19 &  81.27 &  81.18 &  \textbf{\textit{82.20}} &  79.28 &  69.29 &  81.56 \\
 &  Miniboone &  92.68 &  92.27 &  91.10 &  \textit{92.91} &  88.16 &  92.60 &  \textbf{92.91} \\
 &  Occupancy &  99.47 &  \textbf{\textit{99.54}} &  99.45 &  99.50 &  99.33 &  99.40 &  99.33 \\
 &  OpticalDigits &  \textbf{\textit{98.77}} &  94.39 &  98.31 &  98.35 &  92.63 &  94.11 &  96.92 \\
 &  Pageblocks &  95.73 &  \textit{96.97} &  94.03 &  96.84 &  96.53 &  96.89 &  \textbf{97.00} \\
 &  Pendigits &  \textbf{\textit{99.25}} &  95.73 &  98.81 &  98.57 &  88.90 &  98.81 &  98.85 \\
 &  RobotNavigation &  86.14 &  93.97 &  82.24 &  90.25 &  \textbf{\textit{95.20}} &  93.31 &  93.42 \\
 &  SatImage &  \textbf{\textit{90.78}} &  90.32 &  90.66 &  90.77 &  82.30 &  90.75 &  90.46 \\
 &  SocialNetworkAds &  90.47 &  88.75 &  88.48 &  \textbf{\textit{90.72}} &  87.00 &  86.96 &  88.48 \\
 &  Steelfaults &  69.04 &  68.83 &  65.18 &  \textbf{\textit{71.52}} &  58.68 &  68.01 &  71.41 \\
 &  USTaxPayer &  32.08 &  33.67 &  34.25 &  \textbf{\textit{33.86}} &  31.36 &  33.37 &  32.79 \\
\hline
\multirow{17}{*}{\rotatebox[origin=c]{90}{Logistic Regression (LR)}} &  Churn &  80.74 &  81.49 &  80.45 &  81.23 &  82.02 &  73.01 &  82.36 \\
 &  Corel &  31.68 &  \textit{44.36} &  19.32 &  42.36 &  19.55 &  34.80 &  \textbf{48.59} \\
 &  Diabetes &  76.70 &  66.15 &  \textbf{\textit{77.09}} &  75.52 &  65.88 &  62.51 &  76.44 \\
 &  Glass &  57.04 &  \textit{63.60} &  55.99 &  58.34 &  58.84 &  59.44 &  \textbf{63.69} \\
 &  Heart &  82.54 &  81.60 &  81.84 &  \textbf{\textit{84.57}} &  79.99 &  72.95 &  84.22 \\
 &  Miniboone &  \textbf{\textit{91.11}} &  89.91 &  \textbf{\textit{91.11}} &  90.51 &  79.14 &  90.62 &  90.93 \\
 &  Occupancy & \textbf{\textit{99.24}} &  97.98 &  98.56 &  99.05 &  88.59 &  85.42 &  98.80 \\ 
 &  OpticalDigits &  \textbf{\textit{97.06}} &  94.02 &  96.62 &  96.76 &  90.71 &  95.52 &  96.07 \\
 &  Pageblocks &  93.31 &  \textit{96.04} &  91.52 &  95.34 &  94.10 &  96.44 &  \textbf{96.71} \\
 &  Pendigits &  \textit{93.85} &  84.40 &  91.36 &  93.51 &  64.58 &  90.88 &  \textbf{93.97} \\
 &  RobotNavigation &  68.88 &  \textit{86.14} &  65.87 &  77.31 &  55.02 &  80.46 &  \textbf{87.72} \\
 &  SatImage &  84.46 &  81.27 &  \textit{84.52} &  84.30 &  52.86 &  81.93 &  \textbf{85.41} \\
 &  SocialNetworkAds &  82.21 &  79.23 &  \textbf{\textit{83.71}} &  82.73 &  80.49 &  68.02 &  80.24 \\
 &  Steelfaults &  \textit{65.85} &  59.94 &  65.74 &  64.26 &  49.34 &  57.60 &  \textbf{69.36} \\
 &  USTaxPayer &  34.06 &  34.45 &  \textbf{\textit{34.95}} &  33.56 &  34.66 &  34.05 &  33.67 \\
\hline
\end{tabular}
\label{tbl_classify_accuracy}
\end{table}

It is evident from Fig~\ref{fig_classify_corel} and Table \ref{tbl_classify_accuracy} that the accuracies for each classifier vary with different monotonic transformations when data are normalised with the min-max scaler and the original input data (${\bf x}$) did not always produce the best accuracy (e.g., Corel, USTaxPayer). This justifies the argument that classifiers' accuracies vary due to the change in scales of measurement with the traditional min-max normalisation. Also, different classifiers produced best accuracy with different monotonic transformations. This results suggest that a scale appropriate for one classifier may not be appropriate for another. With the mix-max normalisation, data need to be transformed into an appropriate scale that suits the best for the classifier at hand.

In contrast, rank and ARES transformations are robust to such changes in the scales of measurement. ARES transformation almost always produced better accuracy than the traditional rank transformation (except in the USTaxPayer data set with Logistic Regression). This could be because of the uniform distribution of the resulting data in the case of rank transformation. Rank transformation performed really poorly in some data sets (e.g., Corel, OpticalDigits, Pendigits, RobotNavigation, SatImage, Steelfaults etc.) with ANN. ARES transformation produced more consistent and better or competitive to the best classification accuracies across different classifiers and data sets. 

\subsection{Anomaly detection}

We compared the performance of the three contending pre-processing techniques in 8 publicly available data sets listed in Table \ref{tbl_ad_datasets} using two widely used anomaly detectors - Isolation Forest (IF) \cite{IForest_Liu2008} and Local-Outlier Factor (LOF) \cite{LOF_Breunig2000}. After data pre-processing, instances in a data set where ranked based on their outlier scores using IF and LOF and the Area Under the Receiver Operating Curve (AUC) is reported as a performance measure. We used the implementations of IF and LOF provided in Scikit learn with the default parameter settings except `\textit{n\_neighbors}' in LOF which was set to $\lceil\sqrt{N}\rceil$ as recommended by \cite{LOF_Breunig2000}.

\begin{table}[t]
\scriptsize
\centering
\caption{Characteristics of data sets used in the anomaly detection task.}
\begin{tabular}{l @{\hspace{15pt}} r @{\hspace{15pt}} r @{\hspace{15pt}} r}
\hline
Name & \#Features ($M$) & \#Instances ($N$)  & \#Anomalies \\
\hline
Annthyroid & 22 & 7200 & 534  \\
Breastw & 10 & 683 & 239  \\
Ionosphere & 33 & 351 & 126  \\
Lymph & 39 & 148 & 6   \\
Mnist & 97 & 20444 & 676  \\
Pima & 9 & 768 & 268 \\
Satellite & 37 & 6435 & 2036  \\
Shuttle & 10 & 49097 & 3511 \\
\hline
\end{tabular}
\label{tbl_ad_datasets}
\end{table} 

\begin{figure}[t]
\centering
\subfloat[IF]{\includegraphics[width=0.33\textwidth]{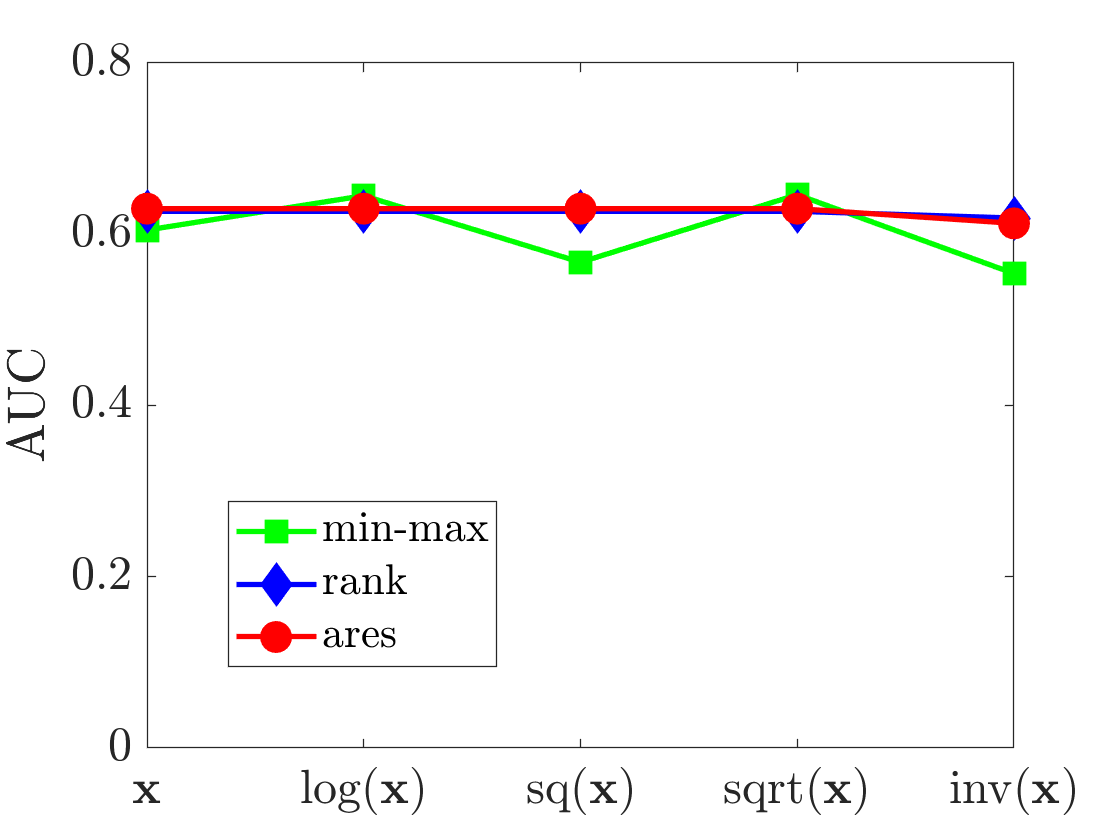}}
\hspace{25pt}
\subfloat[LOF]{\includegraphics[width=0.33\textwidth]{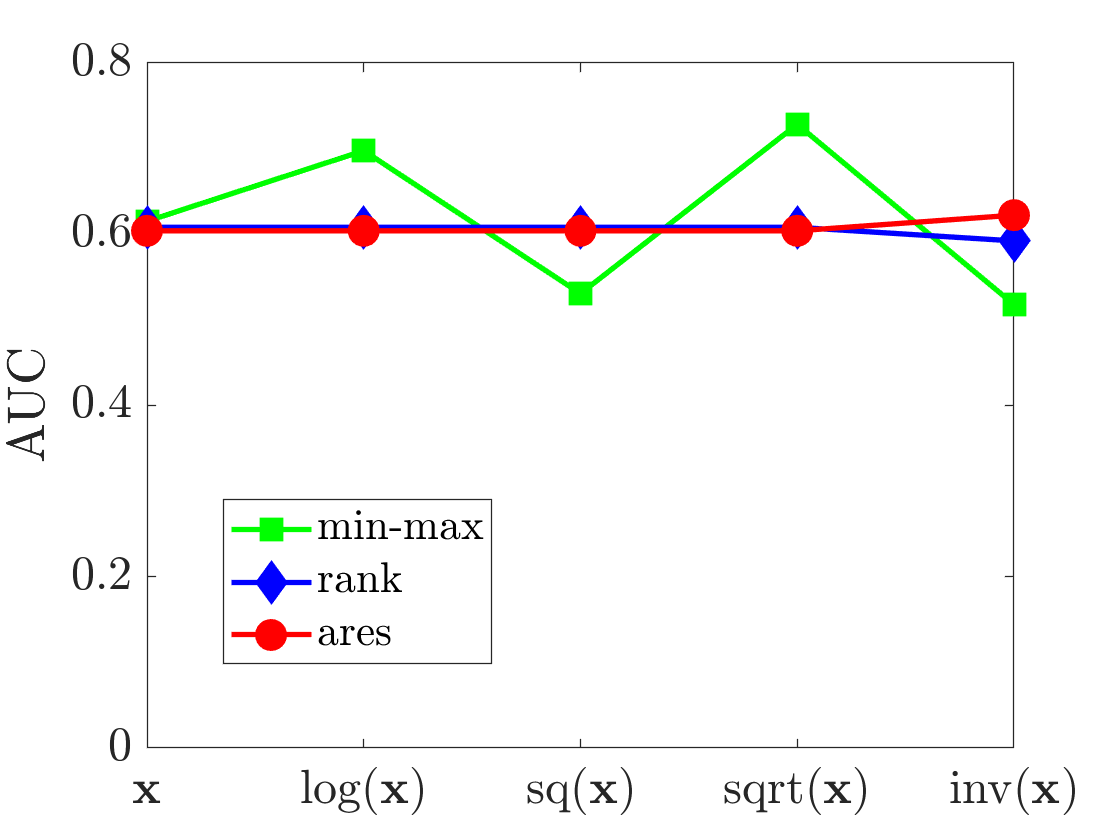}}
\caption{Anomaly detection AUC in the Annthyroid data set.
}
\label{fig_auc_ann}
\end{figure}

We observed the similar trend as in the classification task. The AUC of both IF and LOF varied with different monotonic transformations in the case of min-max normalisation. Unlike min-max normalisation, both rank and ARES transformations produced consistent AUC with different monotonic transformations. Their corresponding results in the Annthyroid data set is presented in Figure \ref{fig_auc_ann}. AUCs of both IF and LOF in all data sets across all the settings of pre-processings --- min-max (original scale and its four monotonic transformations), rank (original scale) and ARES (original scale), are listed in Table \ref{tbl_ad_accuracy}. With min-max normalisation, different monotonic transformations produced the best AUCs in different data sets for both IF and LOF. ARES produced more consistent AUCs than any other contender across all data sets. It produced better results than the rank transformation in many cases except for IF in Mnist and LOF in Annthyroid and Pima. 

\begin{table}[t]
\scriptsize	
\centering
\caption{Area Under the ROC Curve (AUC) in the anomaly detection task. Overall best accuracy in each row is bold faced and the best accuracy result among monotonic transformations in the case of min-max normalisation is italicised.}
\begin{tabular}{c | l |  c @{\hspace{10pt}} c @{\hspace{10pt}} c @{\hspace{10pt}} c @{\hspace{10pt}} c @{\hspace{10pt}} | c @{\hspace{5pt}} | c }
\hline
\multirow{2}{*}{$\mbox{\hspace{15pt}}$} & \multirow{2}{*}{Data set} & \multicolumn{5}{c|}{min-max} & rank & ares \\
\cline{3-7}
&  & ${\bf x}$ & log(${\bf x}$) & sq(${\bf x}$) & sqrt(${\bf x}$) & inv(${\bf x}$) & ${\bf x}$ & ${\bf x}$ \\
\hline
\multirow{8}{*}{\rotatebox[origin=c]{90}{\parbox[c]{2.0cm}{Isolation Forest (IF)}}} &  Annthyroid &  0.6049 &  0.6450 &  0.5673 &  \textbf{\textit{0.6462}} &  0.5541 &  0.6268 &  0.6299 \\
 &  Breastw &  0.9892 &  0.6255 &  \textbf{\textit{0.9930}} &  0.9450 &  0.4097 &  0.6571 &  0.8306 \\
 &  Ionosphere &  0.8338 &  \textbf{\textit{0.8711}} &  0.8278 &  0.8590 &  0.8710 &  0.8587 &  0.8701 \\
 &  Lymph &  \textbf{\textit{1.0000}} &  \textbf{\textit{1.0000}} &  \textbf{\textit{1.0000}} &  \textbf{\textit{1.0000}} &  0.9977 &  0.7441 &  \textbf{1.0000} \\
 &  Mnist &  \textit{0.8351} &  0.8111 &  0.8225 &  0.8295 &  0.7748 &  \textbf{0.8421} &  0.8261 \\
 &  Pima &  0.6650 &  0.4634 &  \textbf{\textit{0.7029}} &  0.6171 &  0.3910 &  0.5975 &  0.6109 \\
 &  Satellite &  0.6585 &  \textbf{\textit{0.7574}} &  0.6909 &  0.7232 &  0.7506 &  0.6572 &  0.6505 \\
 &  Shuttle &  \textbf{\textit{0.9971}} &  0.9967 &  0.9962 &  \textbf{\textit{0.9971}} &  0.9968 &  0.9848 &  0.9445 \\
\hline
\multirow{8}{*}{\rotatebox[origin=c]{90}{\parbox[l]{2.0cm}{Local-Outlier Factor (LOF)}}} &  Annthyroid &  0.6152 &  0.6976 &  0.5307 &  \textbf{\textit{0.7284}} &  0.5179 &  0.6081 &  0.6040 \\
 &  Breastw &  0.3868 &  0.3811 &  0.4106 &  0.3626 &  \textit{0.4136} &  0.4780 &  \textbf{0.5123} \\
 &  Ionosphere &  0.8658 &  0.8662 &  0.8518 &  \textit{0.8796} &  0.7913 &  0.8841 &  \textbf{0.8916} \\
 &  Lymph &  \textbf{\textit{0.9988}} &  0.9965 &  0.9977 &  \textbf{\textit{0.9988}} &  0.9965 &  0.7758 &  \textbf{0.9988} \\
 &  Mnist &  \textbf{\textit{0.8725}} &  0.8096 &  0.8580 &  0.8705 &  0.6046 &  0.8463 &  0.8466 \\
 &  Pima &  \textbf{\textit{0.6189}} &  0.4217 &  0.6059 &  0.5648 &  0.3770 &  0.5532 &  0.5412 \\
 &  Satellite &  0.5685 &  0.5956 &  0.5606 &  0.5864 &  \textbf{\textit{0.6532}} &  0.6239 &  0.6219 \\
 &  Shuttle &  0.5085 &  \textit{0.5359} &  0.5322 &  0.5068 &  0.5337 &  0.5163 &  \textbf{0.5852} \\
\hline
\end{tabular}
\label{tbl_ad_accuracy}
\end{table}


\section{Conclusions and future work}
\label{sec_con}

Many existing data mining algorithms use feature values directly in their model making them sensitive to units/scales used to measure data. Pre-processing of data based on rank transformation has been suggested as a potential solution to overcome this issue. Because the resulting data after pre-processing is uniformly distributed, it may not be very useful in many data mining tasks.

To address the above mentioned issue of the traditional rank transformation using the entire data set, we propose to use ranks in multiple sub-samples of data and then aggregate them. The proposed approach preserves the differences in data values in the given original distribution better than the traditional rank transformation. Our results in classification and anomaly detection tasks using widely used data mining algorithms show that data pre-processing using the proposed approach produces better results than the pre-processing using the traditional rank transformation. In comparison to the most widely used min-max normalisation based pre-processing, it produces more consistent results with the change in units/scales of data. Its results are either better or at least competitive to the min-max normalisation of the resulting data after the best scale mimicked by the monotonic transformations considered in this paper.

Therefore, it is recommended to pre-process the given data using the proposed approach for a good and consistent results of existing data mining algorithms particularly when there is uncertainty on how data are measured or presented.

In future, we would like to test the effectiveness of the proposed data pre-processing technique in: (i) other data mining tasks such as clustering, (ii) application areas such as IoT, sensor networks and health care where data could be in different units/scales as they are measured by different sensors/devices.




\bibliographystyle{splncs}
\bibliography{EnsRank.bib}

\end{document}